# Stateful Evidence-Driven Retrieval-Augmented Generation with Iterative Reasoning

Qi Dong*, Ziheng Lin, Ning Ding

Vocalbeats.AI, 108 Robinson Road, Singapore 068900

**ABSTRACT**

Retrieval-Augmented Generation (RAG) grounds Large Language Models (LLMs) in external knowledge but often suffers from flat context representations and stateless retrieval, leading to unstable performance. We propose Stateful Evidence-Driven RAG with Iterative Reasoning, a framework that models question answering as a progressive evidence accumulation process. Retrieved documents are converted into structured reasoning units with explicit relevance and confidence signals and maintained in a persistent evidence pool capturing both supportive and non-supportive information. The framework performs evidence-driven deficiency analysis to identify gaps and conflicts and iteratively refines queries to guide subsequent retrieval. This iterative reasoning process enables stable evidence aggregation and improves robustness to noisy retrieval. Experiments on multiple question answering benchmarks demonstrate consistent improvements over standard RAG and multi-step baselines, while effectively accumulating high-quality evidence and maintaining stable performance under substantial retrieval noise.

## 1. INTRODUCTION

Large Language Models (LLMs) have achieved strong performance on knowledge-intensive tasks, yet they remain fundamentally limited by parametric memory. They frequently hallucinate, struggle with long-tail knowledge, and exhibit unstable multi-hop reasoning when required to integrate multiple pieces of information [1].

Retrieval-Augmented Generation (RAG) mitigates these issues by grounding generation on external data sources [2, 3]. A retriever first selects relevant documents, and a generator conditions on them to produce answers. While this paradigm improves factuality, most RAG approaches follow a retrieve–then–generate or loosely coupled iterative design. Representative methods enhance retrieval granularity, decompose complex queries, interleave retrieval with chain-of-thought reasoning, or adaptively trigger additional retrieval steps [4]. Despite substantial progress, existing RAG methods remain fundamentally constrained by three structural weaknesses: they treat retrieved documents as flat, unstructured context without explicitly modeling evidence roles; they perform multi-step retrieval in a stateless manner, lacking a persistent reasoning state that stabilizes intermediate conclusions; and they exhibit limited robustness to retrieval noise, as irrelevant or partially misleading documents are neither explicitly filtered nor incorporated as structured constraints [5, 6]. As a result, reasoning quality deteriorates when evidence is incomplete, conflicting, or noisy, revealing a deeper limitation in how current RAG approaches represent and evolve knowledge during inference [7].

In this paper, we address these limitations by reframing the RAG framework as a stateful evidence evolution process built upon a persistent contrastive evidence pool and position our approach within the emerging paradigm of agentic RAG, where autonomous agents iteratively control retrieval, reasoning, and refinement. Instead of directly conditioning on retrieved text, each document is converted into a structured reasoning unit with explicit relevance and confidence signals, decoupling evidence assessment from answer generation. The proposed framework iteratively accumulates supportive and non-supportive evidence into a unified pool, performs evidence-level deficiency analysis to identify informational gaps and conflicts, and updates the query through structured augmentation. These augmented queries actively probe query-related semantic spaces within the corpus to support progressive knowledge accumulation, enabling deeper and more systematic exploration of latent semantic spaces that were not initially retrieved. Through progressive evidence refinement, query updating, and adaptive termination, the framework maintains a stable reasoning state that enables robust multi-hop inference under noisy retrieval conditions. This iterative retrieval–analysis–refinement cycle mirrors how humans search for information: identifying mistakes or missing clues, reformulating queries, and progressively refining supporting information until a reliable conclusion is reached.

*nate.dong@vocalbeats.com

We evaluate our framework on various Question Answering (QA) benchmarks spanning short-form, long-form, and multi-hop reasoning tasks. Across all benchmarks, the proposed framework consistently outperforms conventional retrieve–generate baselines, structured single-step RAG variants [8], and multi-step iterative RAG methods [9, 10]. Ablation studies confirm that (i) structured reasoning units and (ii) contrastive evidence pooling are the key contributors to the performance improvements. To further assess robustness, we inject controlled noise by adding different levels of irrelevant documents into the retrieved results. The framework maintains stable F1 scores as noise levels increase, indicating strong resilience to noisy retrieval conditions.

## 2. RELATED WORKS

The development of RAG has progressed from early retrieve-and-read pipelines toward more adaptive and complex architectures. Our work is motivated by and further expands upon several prominent research directions that have shaped this area.

### 2.1 Standard RAG

RAG aims to enhance LLMs by grounding them in external, non-parametric knowledge sources. Early RAG frameworks typically adopt a two-stage pipeline: a retriever first retrieves documents relevant to the query from a corpus, and a reader or generator, usually an LLM, then produces the final answer conditioned on both the query and the retrieved evidence [2].

This framework helps reduce hallucinations and enables responses based on up-to-date information, alleviating the limitations of static model knowledge. However, single-pass retrieval is mainly suited for single-hop questions and often performs poorly on multi-hop benchmarks such as HotpotQA, which require multi-step reasoning or evidence aggregation across multiple sources [11].

### 2.2 Multi-Step RAG

To overcome the limitations of standard RAG systems, recent work has explored iterative retrieval and reasoning strategies that extend the traditional single-retrieval-then-generation paradigm into multi-step processes. These approaches enable models to progressively gather evidence and refine reasoning across multiple interaction rounds.

Several methods focus on dynamically integrating retrieval with generation. For example, FLARE allows an LLM to decide when additional retrieval is necessary and what information should be fetched during generation [12]. ITER-RETGEN introduces a framework that alternates between retrieval-augmented generation and generation-guided retrieval, demonstrating improvements on multi-hop QA tasks [13]. Similarly, IRCoT incorporates chain-of-thought reasoning to iteratively refine intermediate reasoning steps and guide subsequent retrieval [9].

Other approaches emphasize adaptive retrieval through self-reflection or query reformulation. Self-RAG enables LLMs to retrieve, generate, and critique their outputs through a self-reflective mechanism, improving factual grounding and citation accuracy in open-domain QA and long-form generation [14]. Auto-RAG constructs intermediate retrieval steps using heuristic strategies and answer matching, although its performance still lags state-of-the-art systems [15].

## 3. METHODOLOGY

In this section, we introduce the core algorithm of our RAG framework. The framework is modeled as an iterative, evidence-centric reasoning process, where question answering proceeds through cycles of question decomposition, evidence assessment, error correction, and targeted re-retrieval, continuing until a coherent and well-supported answer is obtained.

### 3.1 Question decomposition

For a given original question $Q$, we design an LLM-based function, denoted as $LLM(Q, p)$, to decompose the question and generate a collection of sub-queries $\{q_1, \dots, q_n\}$ for retrieval. The function takes $Q$ and a natural language instruction prompt $p$ as inputs. The number of generated sub-questions $n$ is not predetermined but generally small, depending on the complexity of $Q$ and the number of distinct reasoning steps or aspects required to address the original question. Decomposing the original question into sub-queries improves the retrieval process by enabling multi-faceted evidence gathering, supporting multi-hop reasoning, and reducing noise from ambiguous queries [16]. For instance, a question such as "What were William Shakespeare's major contributions to literature and his influence on the development of English drama?" involves reasoning over multiple aspects. This question can be decomposed into intermediate sub-queries, such as: (1) *What are the major literary contributions of William Shakespeare?* and (2) *How did William Shakespeare's works influence the evolution of English drama?* where each sub-query corresponds to a distinct reasoning and retrieval step. The original question is retained as the global query and serves as the semantic anchor for subsequent iterative process.

### 3.2 Document retrieval backbone

To retrieve the initial set of top-$k$ candidate documents for each sub-query, we adopt the official DPR dual-encoder checkpoint as the retrieval backbone, without any additional fine-tuning [17]. This choice allows us to decouple retrieval quality from the downstream reasoning process and to focus specifically on analyzing the robustness and performance of the reader component. Notably, this work does not aim to improve retrieval quality itself; instead, it focuses on enhancing the interpretation, filtering, and reasoning over retrieved documents, especially in the presence of noisy or partially irrelevant retrieval results.

### 3.3 Evidence-centric reasoning

#### 3.3.1 Structured reasoning unit

Given a sub-query and a set of retrieved documents $D = \{d_1, \ldots, d_k\}$, we introduce an intermediate representation that transforms each document into a structured reasoning unit (SRU) via a parameterized LLM-based function:

$$SRU = (r, s, e, c) \quad (1)$$

where each component serves a distinct and explicitly defined role. The component $r$ is a discrete relevance decision signal that determines whether the document contributes meaningful information to the query, taking one of three values:

$$r \in \{Supportive, Contextual, Irrelevant\} \quad (2)$$

where Supportive signifies that the document contains information that directly contributes to answering the query; Contextual denotes that the document provides indirect or background information that may assist reasoning but does not directly support an answer; and Irrelevant indicates that the document is uninformative for the query. The component $s$ summarizes the main content of the document for downstream reasoning. The component $e$ captures query-relevant information extracted from the document, or is null if no relevant content is present, and the confidence score $c \in [0,1]$ reflects the estimated reliability of the relevance decision signal $r$. Through this transformation, each retrieved document is converted from long-form text into a structured representation with explicit fields.

For practical implementation and interoperability, each SRU is represented in JSON format, which facilitates machine-readable representation and seamless integration. As a result, downstream reasoning operates over shorter, semantically rich representations, improving overall efficiency and reducing computational cost.

#### 3.3.2 Evidence pool

At the initial iteration, we construct a contrastive evidence pool for the original question. The evidence pool aggregates SRUs derived from all retrieved documents across sub-queries and explicitly retains both positive and negative evidence. Units labeled as Supportive or Contextual are regarded as positive evidence, whereas units labeled as Irrelevant are preserved as negative evidence rather than being discarded. By maintaining both supportive and non-supportive information, this design captures not only what evidence contributes to answering the question, but also what retrieved content is uninformative or potentially misleading, resulting in a stable and explicitly annotated memory of the evidence retrieved during the reasoning process.

Formally, the initial evidence pool is defined as:

$$E^{(0)} = Dedup(\bigcup_{i=1}^{n} \bigcup_{j=1}^{k} SRU_{i,j}^{(0)}) \quad (3)$$

where $SRU_{i,j}^{(0)}$ denotes the structured reasoning unit generated from the *j*-th retrieved document of the *i*-th sub-query at the initial iteration and removes duplicates corresponding to the same source document retrieved by multiple sub-queries. As a result, $E^{(0)}$ provides a unified and contrastive pool of the evidence collected at the initial iteration.

By jointly retaining positive and negative evidence in a single pool, the contrastive evidence pool provides a structured view of both what information supports the question and what information should be avoided. This contrastive formulation prevents premature information loss by exposing negative evidence as explicit constraints rather than discarded noise and establishes a stable foundation for subsequent iterative reasoning and retrieval. As a persistent, evidence-centric memory, $E_0$ provides a unified reasoning context that guides evidence selection, retrieval, and answer synthesis, reduces repeated retrieval and reasoning errors, and supports more reliable multi-step inference in later iterations.

### 3.3.3 Evidence-driven query augmentation

After constructing the contrastive evidence pool from the initial retrieval step, we perform an evidence-driven query augmentation within the same iteration. The purpose of this step is not to generate a new global question, but to derive a compensatory query that remains explicitly grounded in the original question $Q$ and is tailored to address deficiencies revealed by the current evidence.

We design an LLM-based agent to analyze the evidence pool as a unified reasoning state and assess how effectively the retrieved evidence can answer the original question. The agent first evaluates whether the evidence is sufficient to support $Q$; if it is found insufficient, the agent produces one or more types of deficiency signals: **informational gaps**, which indicate under-supported semantic aspects of $Q$; **conflict indicators**, which reflect ambiguities or inconsistencies among supportive evidence; and **negative constraints**, which are abstracted from non-supportive evidence and identify retrieval directions that have proven uninformative. By leveraging the reasoning capabilities of this agent, the analysis captures the evidence pool's limitations in a structured and interpretable manner, providing guidance for subsequent evidence-aware query augmentation.

These deficiency signals serve as the foundation for generating an augmented query $Q'$, which complements the original question. The same LLM-based agent is employed to perform this augmentation, preserving the core intent and semantics of $Q$ while selectively emphasizing under-supported aspects (informational gaps), explicitly prompting clarification for areas of conflicting evidence (conflict indicators), and suppressing retrieval directions associated with negative constraints. As a result, $Q'$ functions as an evidence-aware extension of the original question, rather than an independent or alternative retrieval objective.

Formally, this augmentation process is defined as:

$$Q' = LLM_{agent}(Q \mid E, D\,; task = query\ augmentation)$$
$$D = LLM_{agent}(Q, E; task = deficiency\ analysis) \qquad (4)$$

where $D$ denotes the set of deficiency signals identified from the current contrastive evidence pool $E$. Both deficiency analysis and query augmentation are implemented by the same underlying LLM agent, operating under different reasoning tasks.

### 3.3.4 Iterative evidence accumulation

After completing the initial iteration, the generated augmented query is used as the new input, and the reasoning process is restarted from the decomposition stage. The augmented query is decomposed into a new set of sub-queries, followed by document retrieval and SRU generation using the same procedures as in the initial iteration.

Newly generated SRUs are incrementally integrated into the existing contrastive evidence pool, yielding an updated pool:

$$E^{(t+1)} = E^{(t)} \cup \{SRU_{i,j}^{(t+1)}\}$$

Importantly, the evidence pool is progressively expanding across iterations and serves as a unified reasoning context throughout the iterative process. This persistent accumulation strategy allows evidence obtained in earlier iterations to continuously guide subsequent reasoning steps and reduces redundant retrieval or exploration. The iterative process continues until the accumulated evidence is deemed sufficient to answer the original question or a predefined termination criterion is satisfied. Note that augmented queries are treated as transient control signals and do not persist in the evidence pool, which only stores SRUs of retrieved documents.

## 3.4 Termination strategy

We design a two-stage termination strategy that explicitly distinguishes between cases where sufficient evidence has been accumulated to support answer generation and cases where the question cannot be reliably answered within a bounded number of iterations. All termination decisions, including adaptive early stopping and explicit failure recognition, are performed by the same LLM agent that manages evidence analysis throughout the iterative reasoning process. This design enables the system to adaptively stop when the accumulated evidence is sufficient for reliable answer generation, and to explicitly recognize failure when further retrieval is unlikely to be beneficial. In both cases, termination decisions are made solely based on the evidence collected in the current evidence pool.

### 3.4.1 Adaptive early stopping

At each iteration, after the evidence pool has been expanded with newly retrieved structured evidence and before generating an augmented query, the evidence pool is evaluated with respect to the original question $Q$ to determine whether further retrieval is necessary. Early stopping is triggered when the accumulated evidence is sufficient to support reliable answer generation and the supportive evidence forms a coherent basis for answer synthesis. Under these conditions, further query augmentation is unlikely to improve answer quality, and the iterative reasoning process terminates early, with the final answer generated directly from the current evidence pool.

### 3.4.2 Informed abstention

When the maximum number of iterations is reached and the evidence pool still exhibits unresolved deficiencies, the same LLM agent performs a final evidence-level assessment to determine answerability. Failure is explicitly signaled when key aspects of the original question remain unsupported, newly retrieved evidence is predominantly non-supportive, and negative constraints converge across iterations, indicating diminishing returns from further retrieval. In such cases, the agent clearly identifies that the question cannot be reliably answered based on the available evidence and exercises abstention, effectively knowing what it does not know or cannot answer, rather than generating speculative or hallucinated responses [18, 19].

## 4. EXPERIMENTAL SETTINGS

### 4.1 Benchmarks

We evaluate our framework on benchmarks spanning short-form, long-form, and multi-hop QA tasks to assess its reasoning and evidence integration capabilities. For short-form QA, we use StrategyQA [20], which requires synthesizing factual and commonsense knowledge to produce concise binary answers. For long-form QA, we select ASQA [21] and Natural Questions (NQ) [22], which test the method's ability to gather and combine information from multiple sources to generate detailed explanatory responses. For multi-hop QA, we adopt 2WikiMultiHopQA [23] and HotpotQA [24], which require multi-step reasoning over multiple pieces of evidence or documents to answer complex questions. These benchmarks provide a comprehensive evaluation of the method's capability to filter, reason over, and integrate information across tasks of varying complexity.

### 4.2 Metrics

For evaluation metrics, we report Exact Match (EM), token-level F1 score (F1), and Accuracy (ACC). EM measures whether the generated answer exactly matches the ground-truth answer and is not used for long-form QA, as strict exact matching is overly restrictive for extended responses. F1 computes token-level precision and recall, allowing partial credit for partially correct answers. ACC measures whether the generated answer contains the ground-truth answer string. Together, these metrics provide a balanced evaluation of answer correctness.

### 4.3 Baselines

We evaluate our framework against a diverse set of competitive baselines, including conventional retrieval-augmented generation pipelines, structurally enhanced retrieval methods, and multi-stage retrieval approaches. These baselines represent distinct design paradigms in retrieval granularity, reasoning depth, and knowledge integration strategies.

**LLM**: We include a no-retrieval setting where the LLM directly generates answers without accessing any external knowledge sources. This setting serves as a reference point for assessing the impact of retrieval and external evidence on performance across different QA tasks. **Native RAG**: This baseline follows the conventional retrieval-then-generation paradigm, where a fixed number of documents are retrieved and directly concatenated as input context for answer generation. **Enhanced Single-step RAG**: These approaches improve standard one-shot retrieval by modeling inter-document relations and imposing additional structure over the underlying documents. Representative methods include

RAPTOR [25], which organizes text chunks into a hierarchical summary structure, and HippoRAGv2 [26], which utilizes entity-focused representations to guide retrieval under a single retrieval pass. Such approaches aim to enrich the retrieved context without requiring iterative retrieval. **Multi-step RAG**: Multi-step RAG approaches perform retrieval in multiple stages, gradually incorporating newly obtained information into the answer generation process. IRCoT integrates retrieval within a Chain-of-Thought process [9]. It alternates between natural language reasoning and retrieval steps, treating intermediate reasoning outputs as prompts for fetching additional documents. Self-RAG enhances the generation pipeline with a learned control component that determines whether further retrieval is beneficial at each stage [14].

### 4.4 Implementation details

For all main experiments, we adopt GPT-4.1-mini as the backbone LLM across all compared methods for fair and consistent evaluation. All methods employ the official DPR dual-encoder checkpoint as the retrieval backbone without additional fine-tuning. Retrieval is performed over the publicly released DPR Wikipedia corpus [17] using pre-computed passage embeddings and a FAISS index [27]. The retriever returns the top 5 documents for each query. To ensure both statistical reliability and practical feasibility, all evaluations are conducted on a randomly sampled subset of 2000 instances per benchmark. For each benchmark and each evaluation metric, we report the average score computed over the 2000 sampled instances. Our framework runs for up to five and six iterations, respectively, to analyze the impact of iteration depth while balancing evidence coverage and computational cost. To evaluate generalization capability, we additionally apply the proposed framework with LLaMA3.1-8B-Instruct under the same configurations, using a maximum of five iterations.

## 5. EXPERIMENTAL RESULTS

In this section, we systematically evaluate our framework through a comprehensive set of experiments.

### 5.1 Main results

Table 1. Performance comparison between the proposed framework and representative baselines on five QA benchmarks

| Methods | Benchmarks | | | | | | | | | | | |
|---|---|---|---|---|---|---|---|---|---|---|---|---|
| | StrategyQA | | | ASQA | | NQ | | 2WikiMultiHopQA | | | HotpotQA | | |
| | EM | F1 | ACC | F1 | ACC | F1 | ACC | EM | F1 | ACC | EM | F1 | ACC |
| GPT-4.1-mini | 0.379 | 0.612 | 0.604 | 0.380 | 0.366 | 0.395 | 0.382 | 0.345 | 0.363 | 0.351 | 0.319 | 0.353 | 0.345 |
| Native RAG | 0.427 | 0.645 | 0.638 | 0.433 | 0.413 | 0.429 | 0.403 | 0.397 | 0.420 | 0.404 | 0.371 | 0.411 | 0.419 |
| RAPTOR | 0.476 | 0.660 | 0.659 | 0.446 | 0.429 | 0.450 | 0.439 | 0.446 | 0.459 | 0.462 | 0.422 | 0.452 | 0.468 |
| HippoRAGv2 | 0.480 | 0.671 | 0.643 | 0.460 | 0.439 | 0.452 | 0.433 | 0.437 | 0.460 | 0.458 | 0.435 | 0.461 | 0.453 |
| IRCoT | 0.509 | 0.692 | 0.677 | 0.483 | 0.470 | 0.472 | 0.466 | 0.455 | 0.479 | 0.487 | 0.461 | 0.477 | 0.467 |
| Self-RAG | 0.497 | 0.709 | 0.691 | 0.474 | 0.461 | 0.478 | 0.479 | 0.451 | 0.503 | 0.513 | 0.470 | 0.503 | 0.505 |
| Ours (5 iterations) | **0.533** | **0.728** | 0.717 | **0.511** | 0.495 | **0.551** | 0.536 | **0.508** | 0.642 | **0.651** | **0.511** | **0.613** | 0.608 |
| Ours (6 iterations) | 0.529 | 0.726 | **0.720** | 0.502 | **0.499** | 0.539 | **0.537** | 0.503 | **0.644** | 0.645 | 0.509 | 0.609 | **0.610** |
| Ours (LLaMA3.1) | 0.493 | 0.696 | 0.680 | 0.449 | 0.436 | 0.470 | 0.473 | 0.458 | 0.532 | 0.529 | 0.450 | 0.489 | 0.501 |

The overall performance and comparison results are presented in Table 1, where the best-performing method for each metric is highlighted in bold. Our proposed framework consistently achieves the strongest performance across all benchmarks and evaluation metrics. Performance improvements are especially pronounced on complex multi-hop QA tasks, indicating enhanced cross-document evidence aggregation and compositional reasoning capability. These gains are primarily driven by the SRU-based contrastive evidence pool, which persistently accumulates and refines structured supportive and negative evidence across iterations, enabling stable, cumulative, and adaptive evidence-driven reasoning.

Increasing the number of iterations from five to six yields only marginal improvements on certain benchmarks, suggesting diminishing returns beyond five iterations and indicating that most useful evidence is already captured within five reasoning cycles. When replacing GPT-4.1-mini with the smaller LLaMA3.1-8B-Instruct backbone, overall performance decreases; however, the proposed framework still achieves results comparable to existing multi-step RAG methods, demonstrating strong robustness across backbone models and limited dependence on a highly capable LLM.

### 5.2 Ablation study

To investigate the contribution of the core components in our framework, we conduct a targeted ablation study focusing on two key design elements: (1) the SRU representation, and (2) the explicit preservation of negative evidence in the contrastive evidence pool.

**w/o SRU**: In this variant, retrieved documents are directly incorporated into the evidence pool as raw text without being transformed into structured reasoning units. Consequently, no structured fields are generated. **w/o Negative Evidence**: In this variant, SRUs labeled as Irrelevant are excluded from the evidence pool. Only units categorized as Supportive or Contextual are retained and accumulated across iterations.

The results of the ablation experiments are reported in Table 2. To ensure a fair and controlled comparison, all configurations employ the same retriever, the same GPT-4.1-mini backbone model, a maximum of five iterations, identical evaluation metrics, and the same 2000 sampled instances per benchmark as used in the main experiments. Experiments are conducted on the ASQA and 2WikiMultiHopQA benchmarks.

Table 2. Ablation studies of the proposed framework

| Methods | Benchmarks | | | | |
| --- | --- | --- | --- | --- | --- |
| | ASQA | | 2WikiMultiHopQA | | |
| | F1 | ACC | EM | F1 | ACC |
| Ours (5 iterations) | 0.511 | 0.495 | 0.508 | 0.642 | 0.651 |
| w/o SRU | 0.398 | 0.353 | 0.466 | 0.572 | 0.569 |
| w/o Negative Evidence | 0.487 | 0.450 | 0.471 | 0.611 | 0.606 |

As shown in Table 2, removing either component consistently degrades performance on both benchmarks, confirming the effectiveness of each design element. The w/o SRU variant shows the largest drop, particularly on the long-form QA benchmark ASQA. This substantial decline indicates that SRUs are critical for organizing complex evidence and stabilizing long-form answer generation.

### 5.3 Evidence Accumulation Analysis

To examine whether the proposed iterative reasoning method progressively improves evidence retrieval quality, we analyze the evolution of the evidence pool across iterations. Specifically, at each iteration, we compute the proportion of evidence units labeled as *Supportive* within the cumulative evidence pool, reflecting how much of the stored evidence directly contributes to answering the original question.

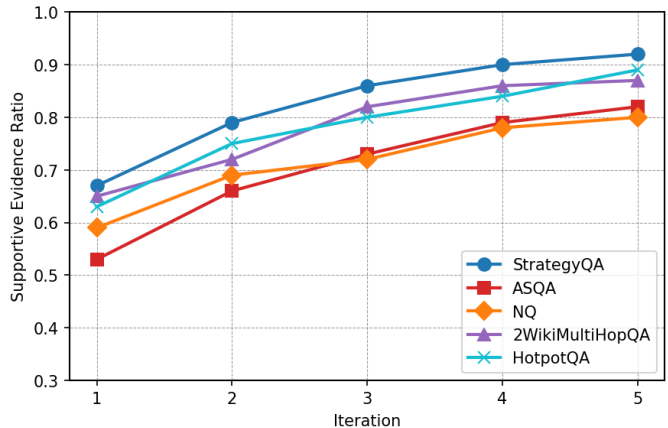

Figure 1. Evolution of supportive evidence ratio across iterations

Using the same 2000 sampled instances per benchmark as in the main results, we compute, for each benchmark and each iteration (1–5), the average Supportive Evidence Ratio across all instances, and present the results in Figure 1.

Across all benchmarks, the Supportive Evidence Ratio increases monotonically over iterations, with gains becoming marginal around the 4th and 5th iterations, indicating a saturation effect where the evidence pool approaches a stable and high-quality state. Simple fact-based benchmark such as StrategyQA achieves consistently higher ratios, while long-form and multi-hop benchmarks show relatively lower values due to increased reasoning complexity. Overall, the steady upward trend demonstrates that the proposed iterative reasoning method progressively accumulates more supportive evidence, confirming its effectiveness.

### 5.4 Retrieval Noise Robustness Analysis

To assess robustness to retrieval noise, we inject randomly sampled irrelevant documents into the top 5 retrieved set for each query, constructing noisy settings with approximately 30%, 50%, and 70% noise ratios, where the noise ratio is defined as the proportion of injected irrelevant documents in the final shuffled document set. We report F1 on NQ and HotpotQA benchmarks, keeping all other experimental settings identical to the main experiments for controlled comparison; results under varying noise ratios are shown in Figure 2.

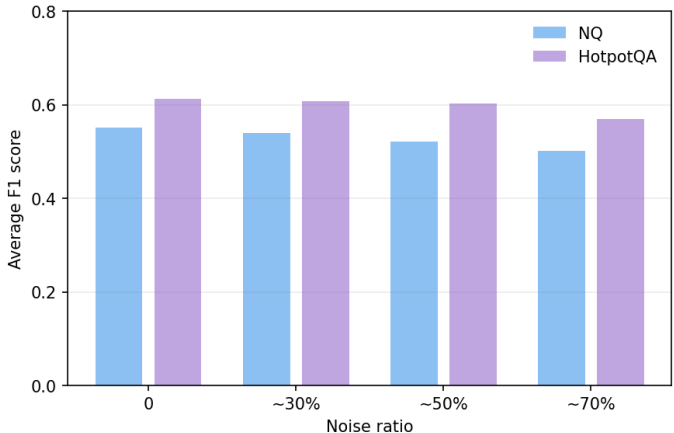

Figure 2. Robustness to increasing retrieval noise

Performance degrades only marginally as noise increases, and even under an extreme noise ratio (e.g., ~70%), the framework still maintains competitive performance. This robustness stems from the SRU-based contrastive evidence pool, which explicitly labels and preserves negative evidence rather than implicitly mixing irrelevant documents into a flat context window. By converting noise into structured negative constraints that guide iterative refinement, the framework sustains stable reasoning under heavy retrieval noise.

## 6. CONCLUSION

In this paper, we propose a stateful evidence-centric RAG framework that improves multi-step reasoning by transforming retrieved documents into SRUs and maintaining a persistent contrastive evidence pool. This design enables explicit modeling of supportive and non-supportive evidence, allowing the framework to perform evidence-driven query augmentation and iterative reasoning under a stable evidence state. Experiments on multiple QA benchmarks demonstrate consistent improvements over conventional RAG pipelines and existing multi-step approaches. Ablation and noise robustness analyses further verify the effectiveness of SRUs and contrastive evidence modeling in improving reasoning reliability. Future work will focus on improving efficiency and extending the framework to larger corpora and broader knowledge-intensive tasks.